\pdfoutput=1

\documentclass[11pt]{article}

\usepackage[]{ACL2023}

\usepackage{times}
\usepackage{latexsym}

\usepackage[T1]{fontenc}

\usepackage[utf8]{inputenc}

\usepackage{microtype}

\usepackage{inconsolata}
\usepackage{booktabs}

\usepackage{xspace}
\usepackage{graphicx}
\usepackage{color}

\newcommand\hiatus{\textsc{hiatus}\xspace}

\newcommand\boardgames{\textit{board}\xspace}
\newcommand\combined{\textit{comb}\xspace}
\newcommand\globalvoices{\textit{global}\xspace}
\newcommand\instructables{\textit{instruct}\xspace}
\newcommand\humanities{\textit{human}\xspace}
\newcommand\stem{\textit{stem}\xspace}
\newcommand\average{avg\xspace}
\newcommand\Reddit{Reddit\xspace}
\newcommand\Realnews{RealNews\xspace}

\newcommand\EightGenres{8 Genres\xspace}
\newcommand\bookcorpus{BookCorpus\xspace}
\newcommand\goodreads{GoodReads\xspace}
\newcommand\amazon{Amazon\xspace}
\newcommand\gmane{Gmane\xspace}
\newcommand\wikiDiscussions{WikiDisc}
\newcommand\pubmed{PubMed\xspace}

\newcommand\sbert{\textsc{sbert}\xspace}
\newcommand\baseSystem{\textsc{Sadiri}\xspace}
\newcommand\randomSystem{\textit{baseline}\xspace}
\newcommand\randomPOSSystem{\textit{baseline-POS}\xspace}

\title{Separating Style from Substance: Enhancing 
Cross-Genre Authorship Attribution through Data Selection and Presentation}

\author{Steven Fincke and Elizabeth Boschee \\
  Information Sciences Institute, University of Southern California\\
  \{sfincke, boschee\}@isi.edu} 

\begin{document}
\maketitle
\begin{abstract}
The task of deciding whether two documents are written by the same author is challenging for both machines and humans. 
This task is even more challenging when the two documents are written about different topics (e.g.\ baseball vs.\ politics) or in different genres (e.g.\ a blog post vs.\ an academic article). 
For machines, the problem is complicated by the relative lack of real-world training examples that cross the topic boundary
and the vanishing scarcity of cross-genre data.
We propose targeted methods for training data selection and a novel learning curriculum 
that are designed to discourage a model's reliance on topic information for authorship attribution 
and correspondingly force it to incorporate information more robustly indicative of 
style no matter the topic.
These refinements yield a 62.7\% relative improvement in average cross-genre authorship attribution, as well as 16.6\% in the per-genre condition.
\end{abstract}

\section{Introduction}
Automatic authorship analysis systems have made significant advances over the past years.
Various tasks require matching documents from the same author. 
We focus here on the task of authorship attribution, where a set of documents by a single author serves as a \textit{query}, and the goal is to identify additional documents by that same author (the \textit{targets}) from a large document collection (called the \textit{haystack}). 
Variation in topic, genre, and domain increase the difficulty of the attribution task.

We report here on results using the newly created \hiatus Research Set (HRS) authorship test set.
This dataset provides both a \textit{per-genre} condition,
where queries and targets are from the same genre,
as well as a more difficult \textit{cross-genre} setting,
where the queries and targets are from separate genres.
We train on author-labeled data collected from the Internet (e.g.\ Reddit, PubMed, GoodReads), where each training author participates in only one such genre.
In this context, we propose \baseSystem 
(\textbf{S}tylometric \textbf{A}uthorship \textbf{D}iscernment \& \textbf{I}nterpretation 
for \textbf{R}ealistic \textbf{I}nputs)\footnote{\textit{Sadiri} is a Bikol word which can be translated as ``(belonging to) oneself'' or ``own'', representing the distinct core of each author's linguistic style.},
an authorship attribution system designed for robust performance in a wide variety of genres, including the cross-genre setting, despite the typically single-genre nature of available authorship training data.

Since each of our training authors participate in only one data source, often writing on a small set of related topics, we see a significant technical challenge in preventing the model from overly associating author identity with topic, genre, or domain. 
To this end, we select and arrange our training data to reduce \baseSystem's opportunities to form positive associations between authorship and topical consistency and instead to focus it on  ``harder'' examples---specifically defined here as those where the model must learn real stylistic information (not just topic similarity) to succeed. To provide \textit{hard positives}, we select training documents for the \textbf{same author} to be topically \textbf{dissimilar}.
Inversely, to provide \textit{hard negatives}, 
we train \baseSystem to differentiate between \textbf{different authors} within pools of \textbf{similar} documents.
These techniques for data selection and presentation provide significant performance gains over state-of-the-art approaches,
especially in the cross-genre context.

\section{Technical Approach}
\subsection{Model infrastructure} \label{sec:core-model}
The core of the \baseSystem system is a fine-tuned LM that produces a vector for a single document. 
e then apply cosine similarity between these vectors to retrieve same-author pairs.
This approach is inspired by LUAR \cite{Soto2021LearningUA},
but simplified to produce vectors for single documents.
We use \textit{RoBERTa-large} \cite{Liu2019RoBERTaAR} as our base model. Accordingly, we apply the RoBERTa tokenizer to the full text and take the first 512 tokens, reflecting RoBERTa's maximum input length. 
The token sequence is fed into RoBERTa, the output of the final hidden layer is extracted and then the mean is calculated for the full input sequence. The output vector is produced by applying a linear transform to reduce dimensionality by a factor of 2: for RoBERTa-large, the final hidden layer has a width of 768, and the output vector length is 384. 
We measure the distance between vectors with cosine similarity 
and train to minimize Supervised Contrastive Loss \cite{Khosla2020SupervisedCL} with respect to our author labels.

We filter our training corpora to include only documents with more than 350 words (to match the configuration of the HRS test set). We then select exactly two documents per author to include in our training process; how these are selected is described in the next section. 
The results reported throughout this paper are the average of scores obtained from models built with two different randomization seeds (42 and 1234).

We train with NVIDIA RTX A6000, which allow us to accommodate vector pairs for 74 authors in each batch. 
However, we train with 4 GPU nodes in parallel so that the loss and gradients for all 4 batches of 74 are calculated separately but are applied together. 
Each epoch contains all the selected training data. We train for four epochs and select the model that performs best on our held-out validation dataset. 

\subsection{Hard positives}\label{sec:model_hard_positives}
We use two documents per author in training. 
When more than two are available, our baseline approach is to select a document pair at random.
In this work, however, our objective is to encourage robustness particularly in a cross-genre setting. To this end, we select the two most topically distant documents instead, producing a ``hard positive'' training example that will ideally force the model to learn elements of stylistic similarity rather than just topical similarity. We also exclude any authors whose most-distant document pair is still ``too similar''. This results in significantly less training data seen by the model, but (as we will show) leads to noticeably improved performance in both per- and cross-genre settings. 

We use SBERT\footnote{huggingface.co/sentence-transformers/all-mpnet-base-v2} \cite{Reimers2019SentenceBERTSE} cosine distance as a reasonable proxy for topical dissimilarity. 
Specifically, we extract SBERT vectors for all of a given author's documents, calculate the cosine similarities between them, and extract the pair with the lowest similarity. 
If the lowest cosine similarity exceeds a prescribed ceiling value, the pair is ignored, and the author is excluded from training. 
We have settled on 0.2 as our default SBERT cosine similarity ceiling but will discuss the impact of this parameter in Section \ref{sec:hard_positive_ablations}.

\subsection{Hard negatives}  \label{sec:learning_cirriculum}  
\baseSystem training batches each contain 74 authors, i.e.\ 74 document pairs. Each pair is selected to be internally topically dissimilar, as described in the previous section, creating ``hard positive'' examples. However, without further intervention, it is likely that the negative examples derived from each batch (e.g.\ comparing document 1 from author A with document 2 from author B) will still be fairly ``easy''---on average, two randomly chosen documents from large Internet corpora will probably be quite distant (no matter how you calculate distance), and the decision that they are written by different authors will be so easy that the model will not be forced learn features that are effective in more challenging situations.  
To mitigate this, we would like all of the authors in a batch to be similar in some way---making the negative examples harder, and forcing the model to learn better representations to compensate. However, we cannot simply gather 148 similar documents (74*2), as we have already forced our document pairs to include topical dissimilarity! For the same reason, representing individual authors with an average of their document vectors is problematic as the pairs are distant by design. 

We therefore develop an approach where we create clusters of authors where only \textit{one} of their two documents is similar to others in the batch. 
That is, we form clusters in which each author provides one document to the dense center and its respective hard negative which falls in the dispersed outer reaches.\footnote{The distribution resembles a \textit{dandelion} more than a billiard ball.}
We pre-specify the number of clusters per batch as a hyperparameter \textit{C}; we default to five but explore this value in Section \ref{sec:hard_positive_ablations}.
To form clusters, we represent each document with a vector (discussed below) and then perform K-means clustering over these vectors and extract an initial set of centroids, enough for all the batches, i.e. \textit{\# batches * C}. 
In the event that two initial centroids are from the same author, one of the two centroids is shifted to the nearest vector from another author to ensure that no author is associated with multiple centroids.

Clusters are then built out by looping repeatedly over the set of final centroids. At each step, we identify the document closest to the given centroid from the remaining pool of unassigned authors and place that author (and all her training documents) to that cluster. 
For efficiency, we use the \textit{search} operation in FAISS \cite{douze2024faisslibrary} to pre-compute a ranked list of documents closest to each centroid, capping at the first 2,024.
This stage ends when no clusters add any authors because 
1) they have already grown to the maximum size, or 
2) all the pre-calculated lists of closest documents have been exhausted.
The relatively small set of remaining authors are added to clusters with spare capacity.
The centroids themselves are clustered by the same algorithm to group clusters into batches (since one cluster is typically not  enough to fill up a batch of 74 authors). 
This additional step yields more coherent batches. 

One critical question is how we generate the document vectors used for clustering. 
Here, we choose to use the vectors generated by the previous iteration of the model. 
(In the first epoch, we use the vector generated by the \baseSystem architecture using the base language model before any fine-tuning.) 
This technique allows us to address the most confusing examples after each iteration and compensate for omissions (or over-training) from previous passes.

\section{Data, Task, \& Metrics}
\textbf{Test Set.} Our test set is the \hiatus Research Set (HRS), developed by IARPA for Phase 1 of the IARPA \hiatus program and made available publicly for research purposes.
All documents in the HRS contain 350 or more words, covering five genres. Please see Appendix \ref{sec:testDataDetails} for more details on access and corpus composition.
For each of these five genres, two groups of documents are provided. 
The \textit{background} documents were found in preexisting sources. For instance, one genre is board game reviews; for this genre IARPA harvested data from \url{boardmangeek.com}.
In contrast, each \textit{foreground} document was created by a writer working under IARPA's supervision to closely approximate a specified background collection. This was done in order to be able to generate a large pool of documents with cross-genre authorship labels; that is, they had each author create documents for more than one genre. 

\textbf{Training Data.} We train with data from eight sources, including our versions of Reddit, a news source (\Realnews), and our version of Amazon reviews.
Additional details are provided in Appendix \ref{sec:trainingDataDetails}.
None of our data sources overlap with those used to create the HRS.
However, the HRS \globalvoices collection can be considered to be the same genre as \Realnews as both are comprised of full news stories,  but in all other cases, tests also require zero-shot genre transfer. 

\textbf{Task Configuration.} The core task we apply to the HRS is one of ranked retrieval---the goal is to return a ranked list of haystack documents with respect to a query, with the hope that the target documents are at the top that list. 
The \hiatus program provided a script to derive test sets for this task from the HRS. First, it identifies a set of test foreground authors. For each selected foreground author, one or more of the author's documents are used as queries, and the rest serve as targets. The remaining foreground documents and all the background documents complete the haystack.

Because of the way the HRS data is constructed, we can control whether targets are required to come from the same genre as the query (the \textit{per-genre} setting) or whether they are assumed to come from different genres than the query (the \textit{cross-genre} setting).
To support the work described here, we ran the \hiatus script to create five per-genre test sets (one for each of the five HRS genres), and six cross-genre test sets (five each with queries from only a single HRS genre and a sixth test set that has queries from all five HRS genres combined together). Additional details are provided in Appendix \ref{sec:testDataDetails}.

\textbf{Metrics.}
We measure performance with \textit{Success@8}: for a given query, was one or more of the eight highest-scoring documents in the haystack written by the same author as the query, i.e. is it a \textit{target} for that query? (Scores are averaged across all queries in a test set.) 
We choose this metric, in part, because it is the primary metric for the \hiatus program; we also prefer it to recall at eight because it is insensitive to the number of targets per author.\footnote{Early development efforts with HRS indicated that Success@8 seemed to correlate well with alternatives such as Mean Reciprocal Rank (MRR), but we have not performed a rigorous analysis.}

\section{Experiments}\label{sec:experiments}
\subsection{Baseline Systems}
We provide two baseline systems.
The first (\randomSystem) is a version of state-of-the-art LUAR \cite{Soto2021LearningUA} which uses only one segment of a single document as input.
The \randomSystem system, like both LUAR and \baseSystem, uses the core model structure and loss function described in \ref{sec:core-model}. However, \randomSystem uses randomly selected same-author document pairs (instead of the \baseSystem hard positives) and randomly combines authors when forming training batches.

The other baseline (denoted \randomPOSSystem) is inspired by the work of \citet{halvani2020improved}: POS tags are extracted for all words, and content words are substituted with tokens indicating only their POS.\footnote{POS tags are produced with the \textit{en\_core\_web\_sm} model within SpaCy \cite{spacy2}. Special tokens for each content POS tag were added to the RoBERTa tokenizer.} 
The goal of this masking strategy is to enhance robustness to topic variation. All other details are the same as for \randomSystem.

In all cases, the baseline systems are trained on the same data (and in the same development environment) as the \baseSystem configurations to which they are compared.

\subsection{Experimental Settings} \label{sec:experimental_settings}
\textbf{Single-Genre Training.} We first present results when training with only one data source, i.e.\ only one genre. We selected two data sources on which to perform single-genre training experiments: (1) \Reddit (because of its size and variety) and (2) \Realnews (because of its genre overlap with \globalvoices in the HRS).
\begin{table*}[!h]
\centering
\begin{tabular}{@{}llllllllll@{}}
\toprule\toprule
data & config &  mode & \boardgames & \combined & \globalvoices & \instructables & \humanities & \stem & \average \\ \midrule
Reddit & \baseSystem &  per   & 0.768 &          & 0.553       & \underline{0.618}   & \underline{0.492}        & 0.535  & \underline{0.593} \\ 
Reddit & \randomSystem  & per &  \underline{0.793} &          & 0.402       & 0.542   & 0.479        & \underline{0.556}  & 0.554 \\ 
Reddit & \randomPOSSystem  & per & 0.625 &          & 0.477       & 0.431   & 0.384        & 0.318  & 0.447 
\\ \midrule
Realnews & \baseSystem &  per  &  \underline{0.793} & & \underline{0.640} & 0.465 & 0.393 & 0.460  & 0.550 \\ 
Realnews & \randomSystem  & per &  0.754 & & 0.581 & 0.417 & 0.315 & 0.455 &  0.504 \\
Realnews & \randomPOSSystem  & per &  0.589 & & 0.512 & 0.306 & 0.341 & 0.328 & 0.415 \\ 
\midrule
\EightGenres & \baseSystem &  per  &  \textbf{0.875} &          & \textbf{0.750}       & \textbf{0.604}   & \textbf{0.530}        & \textbf{0.561}  & \textbf{0.664} \\ 
\EightGenres & \randomSystem  & per &  0.847 &          & 0.506       & 0.549   & 0.461        & 0.485  & 0.570 \\
\EightGenres & \randomPOSSystem  & per & 0.658 &          & 0.524       & 0.361   & 0.371        & 0.384  & 0.459 \\ 
\midrule\midrule
Reddit & \baseSystem &  cross   & \underline{\textbf{0.517}} & \underline{0.283} & 0.213 & \underline{\textbf{0.300}} & \underline{0.401}  & \underline{0.330} &  \underline{0.341} \\ 
Reddit & \randomSystem  & cross & 0.363 & 0.158   & 0.142       & 0.145   & 0.280        & 0.250  & 0.223 \\ 
Reddit & \randomPOSSystem  & cross & 0.271 & 0.164   & 0.099       & 0.095   & 0.271        & 0.205  & 0.184  \\ 
\midrule
Realnews & \baseSystem &  cross  &  0.415 & 0.233   & \underline{0.260}       & 0.185   & 0.334        & 0.285  & 0.286 \\ 
Realnews & \randomSystem  & cross &  0.322 & 0.186   & 0.226       & 0.115   & 0.283        & 0.216  & 0.225 \\
Realnews & \randomPOSSystem  & cross &  0.274 & 0.178   & 0.128       & 0.085   & 0.283        & 0.186  & 0.189 \\ 
\midrule
\EightGenres & \baseSystem &  cross  &  \textbf{0.507} & \textbf{0.328}   & \textbf{0.288}       & 0.275   & \textbf{0.434}        & \textbf{0.368}  & \textbf{0.367} \\ 
\EightGenres & \randomSystem  & cross & 0.353 &  0.178   & 0.189       & 0.145   & 0.229        & 0.258  & 0.225 \\
\EightGenres & \randomPOSSystem  & cross & 0.322 & 0.172   & 0.085       & 0.095   & 0.268        & 0.254  & 0.199 \\
\bottomrule\bottomrule
\end{tabular}
\caption{Performance training on Reddit, Realnews, and the 8 Genre collection on HRS (Success@8). Highest overall scores are in bold; the highest scores among the single-genre systems are underlined.}
\label{tab:various_genre_performance}
\end{table*}
Table \ref{tab:various_genre_performance} demonstrates that our \baseSystem configuration performs best in all contexts, comfortably beating the closest baseline (\randomSystem) with a relative improvement of 7.0\% in the per-genre setting and an even more noticeable 52.9\% in the cross-genre setting.
Our baseline with POS masking of content words (\randomPOSSystem) consistently lags behind both \baseSystem and the \randomSystem system.

Unsurprisingly, training on \Realnews is most effective when testing with queries in the \globalvoices HRS genre, which also consists of news stories. However, in the more typical case where there is no explicit genre match between training and test, we expect the wide variety of topics represented in the \Reddit data source lead to better performance, which is borne out in the results: the system trained on \Reddit does indeed on average out-perform the system trained on \Realnews, by 7.8\% in the per-genre setting and by 19.2\% in the cross-genre setting.

\textbf{Multi-genre training.} 
In machine learning, one can hope that expanding one's training set will increase performance.
However, expanding training to additional domains and genres does not always yield improvements for authorship attribution. 
\citet{Soto2021LearningUA} report modest gains for LUAR when augmenting with their Reddit collection: 4.5\% relative gain in MRR for Amazon and 12.4\% relative for \textit{fanfiction}, but neither Amazon nor \textit{fanfiction} were helpful supplements.

We find similar results for our two baseline systems, as shown in Table \ref{tab:various_genre_performance}, where we compare models trained on one genre (\Reddit or \Realnews) with those trained on eight genres (including \Reddit and \Realnews). 
Adding multi-genre data improves \randomSystem per-genre performance by only a small amount (0.554 to 0.570), and cross-genre performance is essentially unchanged. For \randomPOSSystem, adding multi-genre data actually degrades scores, on average. 
In contrast, for \baseSystem, expanding to \EightGenres significantly improves per-genre performance with performance in the cross-genre context increasing more modestly. The results clearly demonstrate that \baseSystem is much better equipped to capitalize when expanding the range of genres in training, especially in the per-genre condition.

\subsection{Ablation Studies} 
\subsubsection{Hard-positive Ablations} \label{sec:hard_positive_ablations}
A key feature of our approach is how we select documents attributed to the same author for training: we select the pair of documents with the lowest \sbert cosine similarity, and exclude pairs whose similarity exceeds 0.2. We consider two ablations here: 1) varying the ceiling \sbert cosine similarity value, and 2) 
choosing document pairs randomly, including all available authors. 
Table \ref{tab:eight_genres_vary_ceiling} provides results when training on all eight genres.
\begin{table*}[t]
\centering
\begin{tabular}{@{}cccllllllll@{}}
\toprule\toprule
pairs & \# auth &  mode & \boardgames & \combined & \globalvoices & \instructables & \humanities & \stem & \average \\ \midrule

<=0.1 &   129,725 & per   & 0.840 &          & 0.739       & 0.618   & 0.509        & 0.541  & 0.649 \\ 
<=0.2  &  201,673 & per   & 0.875 &          & 0.750       & 0.604   & \underline{\textbf{0.530}}        & 0.561  & 0.664 \\
<=0.4   & 311,563 & per   & \underline{\textbf{0.911}} &          & \underline{0.750}       & \underline{0.708}   & \underline{\textbf{0.530}}        & \underline{0.591}  & \underline{0.698} \\ 
<=0.6  &  368,431 & per   & 0.907 &          & 0.733       & \underline{0.708}   & 0.517        & 0.576  & 0.688 \\ 
<=0.9  & 401,985 & per   & 0.893 &          & 0.687       & 0.639   & 0.487        & 0.546  & 0.650 \\ \midrule
random & 413,169 & per    & \textbf{0.911} &          & \textbf{0.762}       & \textbf{0.771}   & 0.513        & \textbf{0.611}  & \textbf{0.713} \\   

\midrule\midrule
<=0.1  &   129,725 & cross   & \underline{\textbf{0.524}} & \underline{\textbf{0.332}}   & 0.288       & \underline{\textbf{0.320}}   & 0.428        & 0.356  & \underline{\textbf{0.375}} \\ 
<=0.2  &  201,673 & cross   & 0.507 & 0.328   & 0.288       & 0.275   & \underline{\textbf{0.434}}        & 0.368  & 0.367 \\
<=0.4  & 311,563 & cross  & 0.500 & 0.295   & 0.302       & 0.245   & 0.383        & 0.402  & 0.354 \\ 
<=0.6   &  368,431 & cross  & 0.497 & 0.247   & 0.250       & 0.190   & 0.346        & \underline{\textbf{0.409}}  & 0.323 \\ 
<=0.9   & 401,985 & cross  & 0.490 & 0.245   & \underline{\textbf{0.307}}       & 0.255   & 0.350        & 0.379  & 0.337 \\ \midrule
random & 413,169 & cross    & 0.459 & 0.216   & 0.241       & 0.250   & 0.328        & 0.375  & 0.311 \\ 

\bottomrule\bottomrule
\end{tabular}
\caption{Performance training on 8 Genres according to the selection of same-author document pairs for HRS (Success@8). The highest scores in each condition are in bold; the best score among the hard positive configurations is underlined.}
\label{tab:eight_genres_vary_ceiling}
\end{table*}
As we vary the \sbert ceiling from 0.1 to 0.9, we observe a broad trend in declining cross-genre performance; indeed, including increasingly similar document pairs reduces performance in the cross-genre condition on average. 
To a lesser extent, raising the ceiling improves per-genre performance, but the lowest performance is for <=0.1, <=0.6 is somewhat worse than <=0.4, and average performance for <=0.9 is similar to <=0.1.
We chose <=0.2 for our \baseSystem configuration as it provides a compromise between per- and cross-genre performance.

We also consider randomly selecting pairs of documents from all available authors.
Table \ref{tab:eight_genres_vary_ceiling} indicates that this configuration provides the highest per-genre performance, but the lowest (average) cross-genre performance. 
Unlike the \randomSystem systems presented in Tables \ref{tab:various_genre_performance}, 
hard negatives are 
still in place.
In this context, we see that hard positives benefit cross-genre at the expense of the per-genre performance.

\subsubsection{Hard-negative ablations} \label{sec:hardBatchingAblations}
To better understand the impact of the \baseSystem approach to creating training batches (i.e.\ prioritizing ``hard negatives''), 
Table \ref{tab:hard_batching_removal} shows results with and without this technique in place. (Note that in both cases the system assumes our standard ``hard positive'' configuration.)
\begin{table*}[!t]
\centering
\begin{tabular}{@{}ccclllllll@{}}
\toprule\toprule
data & batching & mode & \boardgames & \combined & \globalvoices & \instructables & \humanities & \stem & \average \\ \midrule
\Reddit & hard & per & 0.768 & & 0.553 & \underline{\textbf{0.618}} & 0.492 & \underline{0.535} & \underline{0.593} \\
\Reddit & random & per &  0.747 &  & 0.529 & 0.611 & \underline{0.496} & 0.520 & 0.581 \\ \midrule
\Realnews & hard & per & \underline{0.793} & & \underline{0.640} & 0.465 & 0.393 & 0.460 & 0.550 \\
\Realnews & random & per &  0.714 & & 0.535  & 0.417  & 0.328 & 0.425 & 0.483 \\ \midrule
\EightGenres & hard & per   & \textbf{0.875} &          & \textbf{0.750}       & 0.604   & \textbf{0.530}        & \textbf{0.561}  & \textbf{0.664} \\
\EightGenres & random &   per   &  0.782 & & 0.599 & 0.528  & 0.440  & 0.455  & 0.561  \\ 
\midrule\midrule
\Reddit & hard & cross & \underline{\textbf{0.517}} & \underline{0.283} & 0.213 & \underline{\textbf{0.300}} & 0.401 & \underline{0.330} & \underline{0.341} \\
\Reddit & random & cross &  0.490 & 0.261 & 0.184 & 0.250 & \underline{0.428} & 0.303 & 0.319 \\ \midrule
\Realnews & hard & cross & 0.415 & 0.233 & \underline{0.260} & 0.185 & 0.334 & 0.288 & 0.286 \\
\Realnews & random & cross &  0.367 & 0.221 & 0.217  & 0.130 & 0.304 & 0.273 & 0.252 \\ \midrule
\EightGenres & hard & cross & 0.507 & \textbf{0.328}  & \textbf{0.288}       & 0.275   & \textbf{0.434}        & \textbf{0.368}  & \textbf{0.367} \\
\EightGenres & random & cross   & 0.445 & 0.233 & 0.241 & 0.180 & 0.322 & 0.311  & 0.289 \\ 
\bottomrule\bottomrule
\end{tabular}
\caption{Performance for HRS training on 8 Genres according to batching strategy (Success@8). Top overall performance is indicated with boldface; the best scores for the single-genre configurations are underlined.}
\label{tab:hard_batching_removal}
\end{table*}
We observe initially that the \baseSystem batching approach improves performance in every condition. 
A closer analysis, however, reveals that increasing from just \Reddit to \EightGenres is broadly beneficial \textit{only} with hard negatives in place.
Thus, hard negatives appear to be necessary for achieving the gains reported with multi-genre training in Section \ref{sec:experimental_settings}.

One hyperparameter for the \baseSystem batching approach is the number of document clusters per batch. The results reported in the previous sections target five clusters for each batch of 74 authors. We swept values one through nine: five was a strong choice, but little variance was observed.
As discussed in Section \ref{sec:learning_cirriculum}, we cluster centroids to form batches, and we note that our loss is computed over batches (not clusters). 
This leads us to suspect that clustering the clusters lessens the impact of this hyperparameter and 
causes batches to be largely similar.

\section{Previous work} \label{sec:previousWork}
Work in automatic authorship attribution has previously used information such as character, word, and POS N-grams \cite{Stamatatos2009ASO, DBLP:conf/ifip11-9/StolermanOAG14}, but systems based on pre-trained language models such as BERT \cite{devlin-etal-2019-bert} and RoBERTa \cite{Liu2019RoBERTaAR} have shown greater promise in recent years.
Earlier systems adopted the Siamese network structure employed in this effort but optimized according to a classification loss \cite{Saedi2019SiameseNF, Fabien2020BertAAB}, but more recent systems, e.g. \cite{Soto2021LearningUA, Ibrahim2023EnhancingAV}, made performance gains by switching to Supervised Constrastive Loss \cite{Khosla2020SupervisedCL}.

To limited success, various authorship systems have expanded their architecture to improve robustness to inconsistency in training and test data with regards to topic, genre and domain.
An effort at topic regularization by \citet{Sawatphol2022TopicRegularizedAR} yielded mixed results when training on testing on different sources such as Amazon reviews, Reddit and Fanfiction. 
ContrastDistAA added Mutual Information Maximization to Constrastive learning \cite{10.1145/3589335.3652501} and obtained a 2.1\% relative improvement in F1 on CCAT50, a collection of Reuters articles.
A sequence of papers by \citet{Barlas2021ATL, Barlas2020CrossDomainAA} included the use of an unlabeled \textit{normalization corpus} to enhance cross-topic and cross-genre performance but average accuracy for a BERT-based model falls by 11.9\%, relative, in a cross-topic news article task when the normalization corpus differs from the test set.
Regardless of the dataset, normalization only degrades performance in the cross-fandom environment on PAN18. 

Another common approach to enhancing flexibility is partially obscuring lexical information, focusing on the most distinctive semantic content. 
An approach from \citet{Stamatatos2018MaskingTI} masks all but the most frequent lexical types, retaining varying amounts of detail; 
using an SVM on character N-grams, they observe strong gains in the cross-topic context that largely dissolve cross-genre. 
\citet{halvani2020improved} substituted POS tags for content words for an author verification task.
They compared to a version from \citet{Stamatatos2018MaskingTI} where sequences of letters are masked with a single asterisk (*) for all but the most frequent lexical types,
and reported an average relative gain in accuracy of 3.2\%, with a maximum of 12.5\%.

Like our approach,
\textsc{Valla} \cite{Tyo2022OnTS, Tyo2023VallaSA} seeks hard positives and hard negatives in training.
Their papercompares various approaches in a variety of authorship attribution and verification tasks.
They find that a BERT-based Siamese network trained with constrastive loss performed markedly worse than an N-gram-based approach, especially in cross-topic and cross-genre contexts. 
In an extension of their work, they employ a technique established in computer vision for person re-identification \cite{hermans2017defensetripletlossperson}:
within each randomly selected batch, triplet loss is calculated with the most distant positive and closest negative by Euclidean distance. 
\textsc{Valla} provides a 7.8\% relative increase in accuracy over their baseline with constrastive loss.
They demonstrate the value of calculating loss in training with hard positives and hard negatives, but they only sample within (randomly selected) batches, where \baseSystem mines hard positives and negatives from the full training collection.
\textsc{Valla} shifts to triplet loss in this context while \baseSystem maintains supervised contrastive loss.

\section{Discussion}
\baseSystem employs both hard positives and hard negatives; we consider here how they relate to each other.
First, we observe that hard negatives are consistently helpful, but hard positives bring performance trade offs.
Table \ref{tab:hard_batching_removal} shows that, in the presence of hard positives, hard negatives are beneficial not only in the per-genre condition, but virtually across the board.
However, when we maintain hard negatives, hard positives give higher cross-genre performance at the expense of a loss in the per-genre condition, as shown in Table \ref{tab:eight_genres_vary_ceiling}.

Another asymmetry is that our implementation pursues hard positives and hard negatives by different means. 
We select same-author training pairs by minimizing SBERT cosine similarity; this method selects semantically (and topically) distant pairs, but has not means to directly seek out difference in genre (or subgenre).\footnote{Of course, we have no multi-domain training authors so difference in domain is possible.}
However, hard negatives are identified according to the cosine similarity of the current model.
This discourages associating style with topic, genre and domain to the degree these are encoded by the model at the time.
Our results for random within-genre batching in Section \ref{sec:hardBatchingAblations} suggest that our vectors may encode some genre attributes,  but the present effort will not further probe or characterize our output vectors.
We could impose consistency by detecting hard negatives with SBERT cosine similarity, but early experiments suggested that this method performed worse.
Inversely, we could find hard positives according to \baseSystem vectors, but we have not explored variants along these lines in our present effort. 

\section{Limitations}
We only examine documents with at least than 350 tokens and cannot make any claims about shorter texts.
Also, the HRS test set features elicited \textit{foreground} documents; we would like to evaluate our system on a naturally occurring set of documents, but for the cross-genre condition there are significant practical barriers to amassing such a corpus of any size, as discussed earlier.
We also note that we only consider English-language texts, and performance for other languages has not been addressed. 

\section{Ethics Statement}
The authorship predictions generated by \baseSystem are not always correct: output characterizations of input document collections should be used accordingly.

\section{Conclusion}
We have presented an authorship attribution system that displays remarkable robustness to mixing topics, genres, and domains.
Whereas previous efforts increased modelling machinery in the hope of increasing such functionality, our model architecture is a simple 
RoBERTa-based document encoder trained with supervised contrastive loss.
Instead, our performance gains are attributable to our treatment of our source corpora in training.
Selecting topically dissimilar same-author document pairs is key to improving tests in the cross-genre condition,
Gathering these pairs to establish hard negatives allows us to benefit from large and diverse training corpora and enhances test performance in all conditions, especially the per-genre setting.

\section{Acknowledgements}
This research is supported in part by the Office of the Director of National Intelligence (ODNI), Intelligence Advanced Research Projects Activity (IARPA), via the HIATUS Program contract \#2022-22072200006. The views and conclusions contained herein are those of the authors and should not be interpreted as necessarily representing the official policies, either expressed or implied, of ODNI, IARPA, or the U.S. Government. The U.S. Government is authorized to reproduce and distribute reprints for governmental purposes notwithstanding any copyright annotation therein.

We thank David Jurgens and Jian Zhu for their technical collaboration and extensive data collection efforts on behalf of this work.

\bibliography{anthology,custom}
\bibliographystyle{acl_natbib}

\appendix
\section{Test data details} \label{sec:testDataDetails}
IARPA provides documentation for the Phase 1 HRS data set at https://www.iarpa.gov/research-programs/hiatus;
requests for data for research purposes should be directed to hiatus\_data@umd.edu.
Documents contain a minimum of 350 words, spaning 5 genres, as listed in Table \ref{tab:HRS}.
\begin{table*}[!th]
\centering
\begin{tabular}{@{}lllcc@{}}
\toprule\toprule
name & description &  sources & \# foreground docs & \# background docs \\ \midrule
\boardgames & board game reviews & boardmangeek.com & 212 & 2,432 \\
\globalvoices & citizens journalism & globalvoices.org & 140 & 5,453 \\
\instructables & instructions & instructables.com & 133 & 8,534 \\
\humanities  & literature forums & stackexchange.com & 199 & 9,941 \\
\stem & STEM forums & stackexchange.com & 171 & 11,042 \\
\bottomrule\bottomrule
\end{tabular}
\caption{Composition of the HRS Phase 1 dataset for \textsc{hiatus}.}
\label{tab:HRS}
\end{table*}

A script was provided by the \hiatus program to extract per-genre test sets for the five Phase 1 HRS genres, as well as cross-genre test for each and one combining all five. Details of the version utilized for this paper are provided in Table \ref{tab:HRS_ready}.
\begin{table*}[h]
\centering
\begin{tabular}{@{}lllllll@{}}
\toprule\toprule
 & \multicolumn{3}{c}{per-genre} & \multicolumn{3}{c}{cross-genre}  \\ 
\cmidrule(lr){2-4}\cmidrule(lr){5-7}
 &  \# authors & \multicolumn{2}{l}{mean doc/author} &  \# authors & \multicolumn{2}{l}{mean doc/author}  \\ 
genre &    & queries & targets     &  & queries & targets     \\ \midrule
 
\boardgames & 60 & 1.87 & 2.33   & 64 & 5.72 & 2.28 \\
\combined & & & & 116 & 5.05 & 2.18  \\
\globalvoices & 52 & 1.85 & 1.65  & 49  & 4.57 & 2.16\\
\instructables & 54  & 1.35 & 1.33 & 62  & 6.21  & 1.61 \\
\humanities & 61  & 2.08 & 1.90 & 74 & 5.85 & 2.24 \\
\stem & 52  & 2.12 & 1.90 & 52 & 4.62 & 2.38 \\
\bottomrule\bottomrule
\end{tabular}
\caption{Composition of the HRS Phase 1 test sets with the number of authors and mean number of document per author for 
targets and queries.}
\label{tab:HRS_ready}
\end{table*}
The remainder of the haystack varies modestly in size with over 21k authors and a mean ~2.48 documents per author.

\section{Training data details} \label{sec:trainingDataDetails}
Our training datasets were extracted by our team.
Processing included masking of personally identifying information, e.g. substituting \textit{CREDIT\_CARD} for an actual credit card number, \textit{EMAIL\_ADDRESS} for an email address, etc. 
Details for all eight sources are provided in Table \ref{tab:training_sources}.
\begin{table*}[!h]
\centering
\begin{tabular}{@{}llr@{}}
\toprule\toprule
source &  description & \# docs     \\ \midrule
\Realnews & news stories & 200,602 \\
\bookcorpus & full-length books & 76,529 \\
\Reddit & reddit.com entries & 55,034 \\
\goodreads & book reviews & 32,010 \\
\amazon & reviews on amazon.com & 16,476 \\
\gmane & newsgroups & 16,396 \\
\wikiDiscussions & Wikipedia editorial discussions & 13,868 \\
\pubmed & medical journal articles & 692 \\
\bottomrule\bottomrule
\end{tabular}
\caption{Training data sources}
\label{tab:training_sources}
\end{table*}
\bookcorpus consists of novels and other book-length works which we segmented into multiple \textit{documents} for training corpus; all training \textit{documents} for all other sources are actual full documents. Please note that we amassed our \Reddit and \amazon collections independent of any datasets created for earlier efforts, e.g. \cite{Soto2021LearningUA}.

\end{document}